\documentclass{article}
\usepackage{ijcai26}

\usepackage{times}
\usepackage{soul}
\usepackage{url}
\usepackage[hidelinks]{hyperref}
\usepackage[utf8]{inputenc}
\usepackage[small]{caption}
\usepackage{graphicx}
\usepackage{amsmath}
\usepackage{amsthm}
\usepackage{booktabs}
\usepackage{algorithm}
\usepackage{algorithmic}
\usepackage[switch]{lineno}
\usepackage{multirow}

\title{Investigating the Impact of Histopathological Foundation Models on Regressive Prediction of Homologous Recombination Deficiency
}

\author{
Alexander Blezinger$^1$
\and
Wolfgang Nejdl$^1$\and
Ming Tang$^1$
\affiliations
$^1$Leibniz University Hannover\\
\emails
alexander.blezinger@stud.uni-hannover.de,
nejdl@l3s.de,
tang@l3s.de
}

\begin{document}

\maketitle

\begin{abstract}
    Foundation models pretrained on large-scale histopathology data have found great success in various fields of computational pathology, but their impact on regressive biomarker prediction remains underexplored. In this work, we systematically evaluate histopathological foundation models for regression-based tasks, demonstrated through the prediction of homologous recombination deficiency (HRD) score - a critical biomarker for personalized cancer treatment. Within multiple instance learning frameworks, we extract patch-level features from whole slide images (WSI) using five state-of-the-art foundation models, and evaluate their impact compared to contrastive learning-based features. Models are trained to predict continuous HRD scores based on these extracted features across breast, endometrial, and lung cancer cohorts from two public medical data collections. Extensive experiments demonstrate that models trained on foundation model features consistently outperform the baseline in terms of predictive accuracy and generalization capabilities while exhibiting systematic differences among the foundation models.
    Additionally, we propose a distribution-based upsampling strategy to mitigate target imbalance in these datasets, significantly improving the recall and balanced accuracy for underrepresented but clinically important patient populations. Furthermore, we investigate the impact of different sampling strategies and instance bagsizes by ablation studies.  Our results highlight the benefits of large-scale histopathological pretraining for more precise and transferable regressive biomarker prediction, showcasing its potential to advance AI-driven precision oncology.
\end{abstract}

\section{Introduction}

Foundation models, typically based on transformer architectures and trained in a self-supervised manner on large datasets, learn general-purpose representations that can be adapted to downstream tasks with minimal supervision~\cite{FoundModels}. Recently foundation models have shown great success in various applications of artificial intelligence~\cite{GPT3,Whisper,Climax}. In computational pathology, these models offer particular promise due to the increasing availability of digitized whole slide images (WSIs) from modern slide scanners~\cite{CPathPathahead,PFMSurvey}. WSIs provide rich morphological information to perform analysis which can support tissue classification, prognosis estimation and treatment planning~\cite{CPathAssociation,CpathDefinition,CPathPathahead}.

In histopathology, the transition from discrete classification to regressive biomarker prediction is essential for more nuanced personalized treatment. Predicting continuous biological values allows for a finer assessment of disease state than binary labels. We use Homologous Recombination Deficiency (HRD)—the inability to repair DNA double-strand breaks—as a primary use case for this regressive approach. HRD score is a critical pan-cancer biomarker that predicts patient response to platinum-based chemotherapy and PARP inhibitors~\cite{HRD-Definitions,HRD-Survey}. While it is traditionally assessed by summing three genomic markers (LOH, TAI, and LST), this process is often costly and time-consuming~\cite{DeepSMILE,hoppe2018biomarkers,scarHRD}. Demonstrating the ability to regress HRD scores directly from WSIs offers a faster, more accessible alternative, serving as a benchmark for the broader goals of precision oncology through regressive modeling.

However WSI analysis presents several challenges. Slide images are extremely large, often reaching gigapixel resolution, making end-to-end processing computationally infeasible. Multiple Instance Learning (MIL) addresses this by representing each slide as a bag of image patches ~\cite{TransMil,KatherlabHRD,DeepHRD,ABMIL}, but MIL introduces additional challenges in patch selection, instance aggregation, and interpretability. Furthermore, real-world histopathology datasets are often imbalanced, with clinically relevant conditions being underrepresented~\cite{DataImbalanceComparison,Attention-guidedPrototypemixing}. Models trained on such data can become biased toward majority classes, limiting their ability to detect rare but important cases, a critical concern in medical applications.

In this work, we explore the use of foundation models on regressive biomarker prediction. By addressing the computational challenges of whole-slide images and the clinical reality of data imbalance, our work establishes a benchmark for regressive modeling in precision oncology. Specifically our main contributions are: 


\begin{enumerate}
    \item We provide the first systematic evaluation of histopathology foundation models as feature extractors for regressive HRD prediction. Our results demonstrate that these models consistently outperform baselines in predicting HRD scores across diverse cancer cohorts.
    \item We introduce a novel, distribution-based upsampling algorithm specifically designed to address target imbalance in regression. It significantly enhances the model's predictive power and reliability for underrepresented patient populations, directly addressing a major barrier to the clinical adoption of AI in oncology.
    \item We conduct a comprehensive investigation to identify the effective instance selection and sampling strategies within MIL frameworks.
\end{enumerate}

\section{Related Work} \label{sec:RelatedWork}
\subsection{Self-supervised learning and foundation models in computational pathology}
Self-supervised learning (SSL) is now a key paradigm in computational pathology, enabling large-scale foundation models through representation learning on unlabeled histopathology whole-slide images (WSIs). Many methods originate in computer vision and are adapted to pathology domain. A representative example is RetCCL, which extends the MoCo framework with cluster-guided contrastive learning to treat visually similar WSI patches as positives, thereby improving representations for retrieval and downstream tasks \cite{RetCCL}.


With the increasing availability of large WSI datasets, SSL research shifted toward large-scale pretraining of reusable foundation models, which are often based on transformer architectures. Beyond contrastive approaches, masked image modeling and teacher–student paradigms have gained prominence. Phikon, UNI, and Virchow adopt iBOT- and DINOv2-based pretraining strategies that combine self-distillation with masked image modeling to learn robust patch-level representations~\cite{Phikon,UNI,Virchow,DINOV2}. UNI employs a large Vision Transformer pretrained on 100,000 WSIs, while its successor, UNI-2, significantly scales both model capacity and data, using a ViT-H architecture pretrained on more than 350,000 WSIs~\cite{UNI}. Similarly, Virchow-2 represents one of the largest-scale pathology foundation models to date, trained on over 3 million WSIs with domain-specific adaptations to the DINOv2 framework ~\cite{Virchow2}.


Beyond unimodal vision models, multimodal foundation models have emerged. For instance CONCH aligns histopathology images with captions using the CoCa framework, enabling joint image–text representation learning from over one million paired samples~\cite{CONCH}. To address the challenge that individual foundation models often excel only on specific task categories, the GPFM employs a unified knowledge distillation framework, aligning representations with multiple expert models - including UNI and CONCH - with DINOv2-style self-supervised on WSI patches~\cite{GPFM}. Together, these efforts highlight the rapid progress toward large, diverse, and generalizable pathology foundation models, making the development of large-scale feature extractors an attractive research area.

\subsection{HRD prediction with multiple instance learning}


Predicting Homologous Recombination Deficiency (HRD) from histopathology is typically framed as a weakly supervised problem. Most studies discretize continuous genomic HRD scores into classes using clinically motivated thresholds (e.g., 42) or data-driven percentiles, such as mHRD and tHRD~\cite{SuReTransformer,DeepSMILE}. Methodologically, these approaches rely on embedding-level Multiple Instance Learning (MIL), where a feature extractor processes WSI patches into instance-level features, which are further aggregated into a patient-level  representation of the entire "bag of instances". 
Recently, models such as DeepHRD, DeepSMile, CAMIL regression and SuRe Transformer have demonstrated state of the art performance (Table~\ref{tab:ConclusionResults}) by refining both architectures and training objectives \cite{DeepHRD,DeepSMILE,KatherlabHRD,SuReTransformer}. For instance,  the CAMIL regression model \cite{KatherlabHRD} utilizes RetCCL \cite{RetCCL} for feature extraction and an attention-based MIL (attMIL) framework to predict continuous HRD scores directly rather than discrete labels. Meanwhile, the SuRe Transformer employs clustering-based sampling and radial decay self-attention to efficiently manage the high patch counts inherent in WSIs \cite{SuReTransformer}.



\begin{figure*}[t!] 
    \centering
    \includegraphics[width=\linewidth]{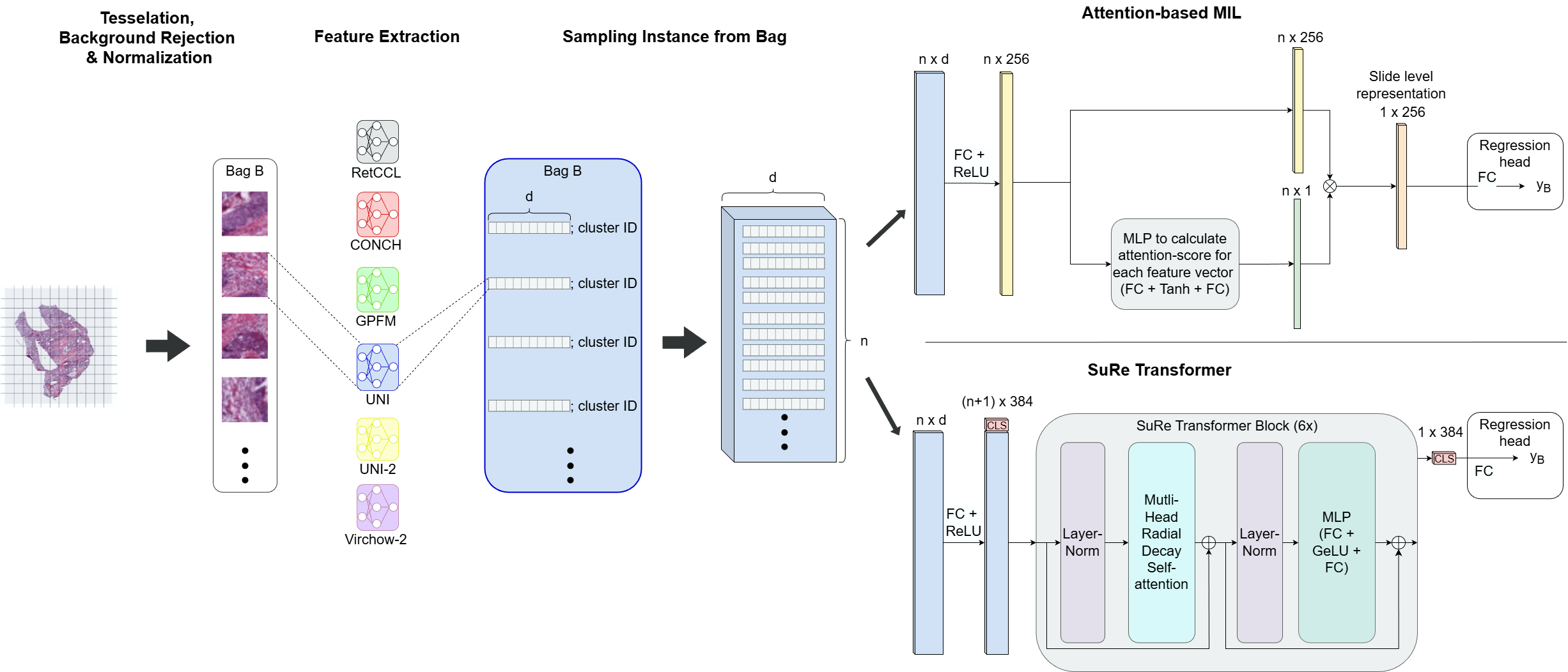}
    \caption{
    The overall architecture of our workflow. WSIs are tessellated and normalized before patches are encoded via feature extraction models to create patient-specific bags of features. From these, $n$ instances of dimension $d$ are sampled (bagsize $n$  x embedding dimension $d$), and processed through one of two aggregation architectures: an attention-based MIL (attMIL) or a SuRe Transformer. The final slide-level representation is passed to a regression head to predict the continuous HRD score ($y_B$).}
    \label{fig:MIL}
    \small 
\end{figure*}

Despite these advancements, most existing models utilize feature extractors that are either pretrained on general images like ImageNet, or on a smaller histopathological datasets like RetCCL~\cite{KatherlabHRD,SuReTransformer,DeepSMILE,DeepHRD}. The potential of modern pathology foundation models, pretrained on millions of WSIs, remains an open research question. This study employs the latest pathology foundation models as feature extractors to improve generalization. Furthermore, by adopting a regressive approach, we aim to provide a more flexible alternative to traditional fixed-threshold classification.

\section{Method}

\subsection{The WSI analysis framework}



Following the the end-to-end WSI preprocessing pipeline \cite{STAMP} which was also used in CAMIL regression model \cite{KatherlabHRD}, WSIs are preprocessed via tessellation, background rejection, and color normalization. For feature extraction, we use RetCCL as our baseline and also implement our own feature extractors with CONCH, GPFM, UNI, UNI-2, and Virchow-2 to evaluate the impact of foundation models. 


To curate patient-level bags, we apply K-Means clustering (k=50) to the extracted features. These bags serve as input for two MIL aggregation architectures: an attention-based model attMIL \cite{KatherlabHRD} and the SuRe Transformer \cite{SuReTransformer}. Both models are trained to predict a continuous HRD value, which is subsequently converted into a binary classification (HRD+/-) in order to compare with previously published results and to further discretize the assessed HRD status.

\subsection{Sampling Strategies}
Transformer-based MIL architectures are computationally sensitive to long input sequences. Since a single WSI can contain tens of thousands of patches, models typically process only a representative subset. To create these training and test instances out of the patient bag of features, three sampling strategies were implemented:

\subsubsection{Cluster Size-Weighted Sampling}

Proposed by \cite{SuReTransformer} for the SuRe Transformer, this is the default strategy for this study. Feature vectors are sampled proportionally to their K-means cluster sizes up to a specified bag size $S$. The number of feature vectors $N_i$ from cluster $i$ is determined by its relative size $C_i$ and the total size of the patients bag $B$:

\begin{equation}
\label{eq:SuRe_Sampling}
N_i = S\times \frac{C_i}{B}
\end{equation} 

The implementation ensures each cluster is represented by at least one vector. For example, if $k=50$ and $S=50$, exactly one vector is drawn per cluster.

\subsubsection{Clustered Random Sampling}
It is a variation of the cluster size-weighted sampling strategy. It still guarantees to sample at least one feature vector per cluster, but instead of sampling amounts based of the relative size of each cluster, it draws a random amount of feature vectors without replacement for each cluster. Thus the sampling amount per cluster $N_i$ is limited by the size of each cluster and the sum of $N_1 + ... + N_k$ is equal to the specified bagsize $S$.

\begin{algorithm}[b!]
    \caption{Distribution-based Upsampling Algorithm}
    \label{alg:Upsampling}
    \textbf{Input}: $binCounts$: histogram counts of continuous target value distribution; $patientBins$: patient bags associated with each bin\\
    \textbf{Parameter}: $sample$: sampling function to create instances; $n\_bins$: number of bins (e.g. 7); $\alpha$: budget cap ratio (e.g. 0.65); $\beta$: budget scaling factor (e.g. 0.25);\\
    \textbf{Output}: $trainingInstances$: upsampled created instances to add to the training dataset \\
    \begin{algorithmic}[1] 
        \STATE Let $trainingInstances = \text{sampled instances from the training dataset}$
        \STATE $maxVal \gets \max(binCounts)$
        \STATE $budgetCap \gets \text{round}(maxVal \times \alpha)$ 
        \FOR {i in binCounts.length}
            \STATE $binBudget \gets \min(budgetCap, (maxVal - binCounts[i]) \times \beta)$
            \FOR {n in binBudget}
                \STATE $bag \gets \text{select random from } patientBins[i]$
                \STATE $trainingInstances.add(sample(bag))$
            \ENDFOR
        \ENDFOR
    \STATE \textbf{return} $trainingInstances$
    \end{algorithmic}
\end{algorithm}

\subsubsection{Random Sampling}
This sampling stratedy draws feature vectors randomly out of the patients bag without replacement up to the specified bagsize $S$. It does not take the k-Means clustering into account.\\

Due to the low complexity of attMIL architecture, our investigation also includes a configuration where every available feature is included -- the bag size for each patient equals the total number of patches extracted from their associated WSIs. This is not possible in SuRe transformer aggregation architecture.

\subsection{Upsampling Algorithm}
To address the data imbalance problem found in our clinical datasets, we implemented an cluster-size-based upsampling algorithm (Algorithm \ref{alg:Upsampling}).  
The purpose of this algorithm is to smooth the distribution by repeatedly sampling image patches from patients with rare HRD values. 
By generating a higher number of non-identical training instances from these rare cases, the model is encouraged to better capture underrepresented HRD values while mitigating the risk of overfitting to specific image patches.

To achieve this, the algorithm first constructs a binned distribution of the target HRD values. Patients are assigned to bins, and a sampling budget is calculated for each bin based on the difference between its size and the size of the largest bin. To preserve the overall characteristics of the original distribution and ensure that upsampling does not distort the underlying data structure, we introduce a budget scaling factor, $\beta$.

The algorithm then samples instances from randomly selected patients according to the computed budget for each bin and appends these newly generated instances to the training dataset. Each new instance consists exclusively of patch features from a single patient to prevent the mixing of data across patients. Furthermore, to prevent excessive resampling from patients in very small bins—thereby reducing the risk of patient-specific overfitting—a budget cap is applied. This cap limits the maximum allowable sampling budget per bin and is defined as a function of the largest bin size, scaled by a factor $\alpha$.



\begin{figure*}[t!] 
    \centering
    \includegraphics[width=\linewidth]{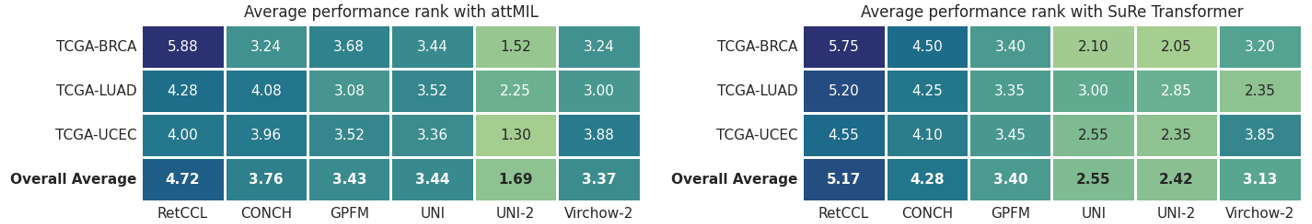}
    \caption{Performance rank of TCGA internal crossvalidation. The rank is computed by comparing the rank of each model on each individual fold.}
    \label{fig:Experiment 1}
\end{figure*}

\section{Experiments}

\subsection{Datasets}
To investigate the impact of the foundation models, experiments on a total of five patient cohorts were conducted. From the TCGA project the Breast Invasive Carcinoma (BRCA) cohort (2,884 WSIs from 1,026 patients), the Uterine Corpus Endometrial Carcinoma (UCEC) cohort (1,224 WSIs from 502 patients), and the Lung Adenocarcinoma (LUAD) cohort (1,397 WSIs from 482 patients) were used for training the prediction models for each cohort. For external validation, an additional LUAD cohort (492 WSIs from 105 patients) and a UCEC cohort (380 WSIs from 98 patients) from the CPTAC project were utilized. Each patient is associated with a specific HRD value calculated using scarHRD \cite{scarHRD} and acquired by \cite{KatherHRDDataSourcePaper}, enabling patient-level predictions across all datasets.

An analysis of the training label distributions within the TCGA cohorts reveals a clear presence of class imbalance. In particular, the BRCA and UCEC cohorts exhibit a pronounced skew toward lower HRD values, making the HRD- class the majority. Specifically, in the BRCA dataset, 2,080 patients belong to the HRD- class compared to 804 patients in the HRD+ class, while in the UCEC dataset, 1,043 patients are HRD- and 181 patients belonging to HRD+. In contrast the TCGA LUAD cohort is more evenly distributed across the classes with 515 HRD+ and 882 HRD- patients. However especially high HRD values still remain rare within this cohort. 


\subsection{Internal TCGA Cohort Validation} \label{tcga_cross}
We conducted five fold cross-validation across three TCGA datasets. MIL models based on the attMIL architecture or the SuRe Transformer were trained and evaluated with the feature vectors from each of the extraction models. To create instances, we utilized a cluster-size weighted sampling strategy with bagsizes of 600, 800, 1000, and 12000. Additionally we evaluated the attMIL architecture using all available features for each patient. This resulted in 54 unique configurations for each cohort: 6 extraction models x (4 bagsizes x 2 architectures + attMIL with all features).  The models were compared using the median AUROC across all folds.\\
\begin{table}[t!]
\centering
\resizebox{\linewidth}{!}{
\begin{tabular}{@{}lrrrrrr@{}}
\toprule
\multirow{2}{*}{Cohort} & \multicolumn{6}{c}{Extraction model}                                             \\ \cmidrule(l){2-7} 
 & \multicolumn{1}{l}{RetCCL} & \multicolumn{1}{l}{CONCH} & \multicolumn{1}{l}{GPFM} & \multicolumn{1}{l}{UNI} & \multicolumn{1}{l}{UNI-2} & \multicolumn{1}{l}{Virchow-2} \\ \midrule
TCGA-BRCA               & 0.7713 & 0.8012 & 0.8078 & \underline{0.8140} & \textbf{0.8304} & 0.8121           \\
TCGA-LUAD               & 0.6740 & 0.6939 & 0.7144 & \textbf{0.7203}  & 0.7191          & \underline{0.7194} \\
TCGA-UCEC               & 0.8070 & 0.8169 & 0.8316 & \underline{0.8412} & \textbf{0.8479} & 0.8206           \\ \bottomrule
\end{tabular}
}
\caption{Internal TCGA cohort validation. The median (5-folds) AUROC from the attMIL and SuRe Transformer architecture are averaged. Bold and underscore represent the best and the second best performance respectively}
\label{tab:R1_AvgAurocs}
\end{table}
Across the five-fold TCGA internal validation, all evaluated foundation models consistently outperformed the RetCCL baseline. Tables\ref{tab:R1_AvgAurocs} shows the average AUROC of attMIL and SuRe Transformer architecture, while Fig.~\ref{fig:Experiment 1} illustrates the performance rankings of each extraction model individually.
Among the foundation models, UNI-2 emerged as the strongest overall feature extractor (Tables\ref{tab:R1_AvgAurocs}). Even in the exception case of TCGA-LUAD where UNI-2 is slightly underperformed than UNI, it still achieved a higher overall performance rank (Fig.~\ref{fig:Experiment 1}). In overall performance, UNI is the second-best with Virchow-2 follows closely. GPFM and CONCH are ranked fourth and fifth respectively. Notably, even the lowest performing foundation model CONCH surpassed the RetCCL baseline across all datasets.


\begin{table}[t!]
\centering
\begin{tabular}{@{}lrrr@{}}
\toprule
\textbf{}        & \multicolumn{3}{c}{\textbf{Utilized Threshold}}                                      \\ \cmidrule(l){2-4} 
\textbf{Publication}      & \multicolumn{1}{r}{\textbf{\textgreater 42}} & \multicolumn{1}{r}{\textbf{mHRD}} & \multicolumn{1}{r}{\textbf{tHRD}} \\ \midrule
CAMIL regression \shortcite{KatherlabHRD} & \multicolumn{1}{r}{0.78} & \multicolumn{1}{r}{\textbf{-}} & \multicolumn{1}{r}{-}    \\
DeepHRD    \shortcite{DeepHRD}      & \multicolumn{1}{r}{0.81} & \multicolumn{1}{r}{\textbf{-}} & \multicolumn{1}{r}{-}    \\
DeepSmile \shortcite{DeepSMILE} & \multicolumn{1}{r}{-}    & \multicolumn{1}{r}{0.75}       & \multicolumn{1}{r}{0.81} \\ 
SuRe Transformer \shortcite{SuReTransformer} & \multicolumn{1}{r}{-}    & \multicolumn{1}{r}{0.81}       & \multicolumn{1}{r}{0.89} \\ \midrule
\textbf{Extraction Model} & \multicolumn{1}{r}{\textbf{}}                & \multicolumn{1}{r}{\textbf{}}     & \multicolumn{1}{r}{\textbf{}}     \\ \midrule
RetCCL                                & 0.7737                   & 0.6841                         & 0.8511                   \\
CONCH                                 & \textbf{0.8159}          & 0.7218                         & 0.8768                   \\
GPFM                                  & 0.8096                   & 0.7190                         & 0.8677                   \\
UNI                                   & \textbf{0.8131}          & 0.7202                         & {\textbf{0.9013}}    \\
UNI-2                                 & {\underline{\textbf{0.8367}}}    & 0.7312                         & {\underline{\textbf{0.9084}}}    \\
Virchow-2                             & \textbf{0.8149}          & 0.7161                         & 0.8686                   \\ \bottomrule
\end{tabular}
\caption{Performance comparison (AUROC) of state-of-the-art literature and our methods on TCGA-BRCA dataset. Here all extraction models were coupled to attMIL . 
To discretize the HRD scores into binary classes, the clinically relevant threshold 42, the data median (mHRD; 27), and tertiles (tHRD; 17, 37) are used. Bold values indicate extraction models that exceed the performance of previously published literature; underscored values identify the overall top-performing model.
}
\label{tab:ConclusionResults}
\end{table}

Overall, foundation-model-based feature extraction substantially improved HRD prediction performance across all evaluated TCGA cohorts. 
These results gain particular significant when compared to state of the art models such as DeepHRD, SuRe Transformer, CAMIL regression and DeepSmile~\cite{KatherlabHRD,DeepSMILE}. When paired with the attMIL architecture on TCGA-BRCA, UNI-2 surpassed existing best results for both the clinically relevant threshold of 42 and the data-percentile driven tHRD.

\subsection{Generalization Across External Cohorts}
To assess generalization performance, we trained our models on the entire TCGA cohorts and evaluated on CPTAC cohorts that are either of the same cancer type or a different cancer type.

\subsubsection{External Validation with the Same Cancer Type}
\begin{figure*}[t!] 
    \centering
    \includegraphics[width=\linewidth]{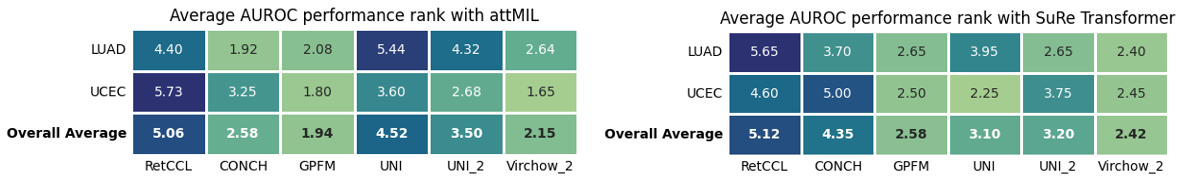}
    \caption{Performance rank for external validation of the same cancer type. The rank is computed by comparing the rank of each model on each individual fold.}
    \label{fig:Experiment 2}
\end{figure*}

We train our models on the TCGA cohorts and evaluate on the corresponding CPTAC cohorts of the same cancer type -- either LUAD or UCEC. The performance was averaged across five random seeds and multiple bag-size configurations, and evaluated by AUROC scores. As shown in Table~\ref{tab:Result_2_Aurocs} and Fig.~\ref{fig:Experiment 2}, foundation models continue to outperform the RetCCL baseline. But within foundation models, there is markedly different rankings compared to the TCGA internal validation experiment. Specifically UNI-2 or UNI no longer shows strong performance while Virchow-2 becomes the best extraction model in both LUAD and UCEC cancer types.

\begin{table}[b!]
\centering
\resizebox{\linewidth}{!}{
\begin{tabular}{@{}lrrrrrr@{}}
\toprule
\multirow{2}{*}{Test Cohort} & \multicolumn{6}{c}{Extraction Model}                                             \\ \cmidrule(l){2-7} 
 & \multicolumn{1}{l}{RetCCL} & \multicolumn{1}{l}{CONCH} & \multicolumn{1}{l}{GPFM} & \multicolumn{1}{l}{UNI} & \multicolumn{1}{l}{UNI-2} & \multicolumn{1}{l}{Virchow-2} \\ \midrule
CPTAC-LUAD              & 0.7649 & \underline{0.8208} & 0.8207           & 0.7699 & 0.8024 & \textbf{0.8219} \\
CPTAC-UCEC              & 0.9503 & 0.9707           & \underline{0.9881} & 0.9839 & 0.9822 & \textbf{0.9885} \\ \bottomrule
\end{tabular}
}
\caption{
External validation with the same cancer type. Models were trained on TCGA cohorts and tested on CPTAC cohorts of the same cancer type. The results are reported as average AUROC.}
\label{tab:Result_2_Aurocs}
\end{table}

\subsubsection{External Validation with a Different Cancer Type}
Previous work has shown that HRD classifiers trained on a single cancer type can partially generalize to other cancer types~\cite{KatherlabHRD,SuReTransformer}. Given that strong generalization is a defining characteristic of foundation models, we evaluated whether foundation model–based feature extractors further improve this cross-cancer-type generalization. 
We conducted two experiments using the attMIL architecture where models were trained on TCGA-LUAD and tested on CPTAC-UCEC and vice versa. We used bag size 600 and 1200, as well as a no-sampling configuration. The results were averaged across five random seeds. As shown in Table~\ref{tab:AUROC_intercohort}, foundation models outperformed the RetCCL baseline most of the times. UNI-2 significantly surpassed other foundation models. Notably the performance gap increased further in the TCGA-UCEC → CPTAC-LUAD direction, where RetCCL shows an extremely low AUROC (0.4599) and UNI-2 has a AUROC of 0.7672, a 40\% increase of performance. 

Overall, these results underscore the superior generalization capabilities of foundation models, which might indicate that foundation models create more "tissue-independent" representations.UNI-2 demonstrated the most consistent and best transfer performance within this experiment. We noticed an exceptional weak performance of UNI in the LUAD → UCEC setting. This suggests limitations in the transferability of UNI-based features when trained on TCGA-LUAD, and this phenomenon should be further investigated with additional datasets.

\begin{table}[t!]
\centering
\resizebox{\linewidth}{!}{
\begin{tabular}{@{}lrrrrrr@{}}
\toprule
\multirow{2}{*}{Train $\rightarrow$ Test} & \multicolumn{6}{c}{Extraction model}                         \\ \cmidrule(l){2-7} 
 & \multicolumn{1}{l}{RetCCL} & \multicolumn{1}{l}{CONCH} & \multicolumn{1}{l}{GPFM} & \multicolumn{1}{l}{UNI} & \multicolumn{1}{l}{UNI-2} & \multicolumn{1}{l}{Virchow-2} \\ \cmidrule(r){1-1}
LUAD $\rightarrow$ UCEC & 0.8933 & 0.9353 & 0.9039 & 0.8163 & \textbf{0.9528} & 0.9020 \\
UCEC $\rightarrow$ LUAD & 0.4599 & 0.4655 & 0.5543 & 0.6198 & \textbf{0.7672} & 0.5462 \\ \bottomrule
\end{tabular}
}
\caption{
External validation with difference cancer types. Models were trained on TCGA cohorts and tested on CPTAC cohorts of different cancer types. Results are reported as average AUROC.}
\label{tab:AUROC_intercohort}
\end{table}

\subsection{Upsampling for Imbalanced Data} \label{upsampling}
We noticed that even our models achieved high AUROC scores on many settings, the performance in terms of balanced accuracy and recall was considerably low.  For example, in the case of external validation with the same cancer type, all extraction models leads to balanced accuracy below 0.69 and recall below 0.52 (top two rows in Table \ref{tab:UpsampledBalAccLUAD}), despite AUROCs are above 0.76 (Table \ref{tab:Result_2_Aurocs}) 
We hypothesize that such discrepancy is largely due to data imbalance and could be mitigated with an appropriate upsampling strategy. 

\begin{figure}[t!] 
    \centering
    \includegraphics[width=130pt]{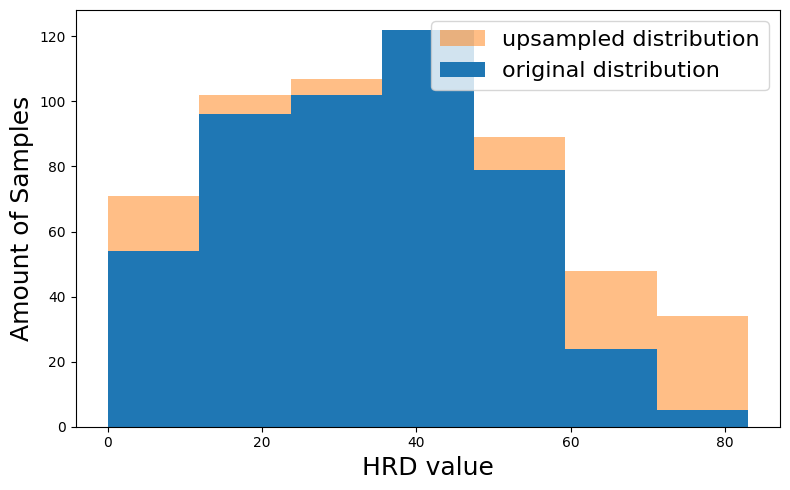}
    \caption{Original vs. upsampled distribution of the TCGA LUAD after applying the distribution-based upsampling algorithm}
    \label{fig:UpsampledDistribution}
\end{figure}

We developed a distribution-based upsampling algorithm and applied to our training data. Figure~\ref{fig:UpsampledDistribution} illustrates the effect of this algorithm on the TCGA-LUAD training cohorts (Alg.~\ref{alg:Upsampling}, with 7 bins, $\alpha=0.65$, $\beta=0.25$). When we apply this upsampled data to our models and evaluate on the external CPTAC cohorts, we observe substantial improvements in recall and balanced accuracy across all extraction models. Table~\ref{tab:UpsampledBalAccLUAD} shows the result in CPTA-LUAD cohort, where balanced accuracy increased by 0.029–0.047 and recall for HRD+ patients improved markedly by 0.081-0.222 depending on the extraction models.

\begin{table*}[h]
\centering
\resizebox{280pt}{!}{
\begin{tabular}{@{}ccrrrrrr@{}}
\toprule
\multirow{2}{*}{\begin{tabular}[c]{@{}c@{}}Training \\ Data Distribution\end{tabular}} &
  \multirow{2}{*}{Metric} &
  \multicolumn{6}{c}{Extraction Model} \\ \cmidrule(l){3-8} 
 &
   &
  \multicolumn{1}{c}{RetCCL} &
  \multicolumn{1}{c}{CONCH} &
  \multicolumn{1}{c}{GPFM} &
  \multicolumn{1}{c}{UNI} &
  \multicolumn{1}{c}{UNI-2} &
  \multicolumn{1}{c}{Virchow-2} \\ \midrule
\multirow{2}{*}{original} &
  Bal. accuracy &
  0.6056 &
  0.6835 &
  0.6797 &
  0.5829 &
  0.6441 &
  0.6894 \\
 &
  Recall HRD + &
  0.3206 &
  0.5048 &
  0.4937 &
  0.2413 &
  0.4206 &
  0.5159 \\ \midrule
\multirow{2}{*}{upsampled} &
  Bal. accuracy &
  0.6346 &
  0.7122 &
  0.7190 &
  0.6296 &
  0.6469 &
  0.7216 \\
 &
  Recall HRD + &
  0.4964 &
  0.7268 &
  0.7071 &
  0.4375 &
  0.5018 &
  0.7054 \\ \midrule
\multirow{2}{*}{\textit{\begin{tabular}[c]{@{}c@{}}Improvement \\ through upsampling\end{tabular}}} &
  \textit{Bal. accuracy} &
  \textit{0.0290} &
  \textit{0.0286} &
  \textit{0.0393} &
  \textit{0.0467} &
  \textit{0.0028} &
  \textit{0.0322} \\
 &
  \textit{Recall HRD +} &
  \textit{0.1758} &
  \textit{0.2220} &
  \textit{0.2135} &
  \textit{0.1962} &
  \textit{0.0812} &
  \textit{0.1895} \\ \bottomrule
\end{tabular}
}
\caption{Average balanced accuracy and recall of HRD+ on CPTAC-LUAD when trained on original or upsampled TCGA-LUAD}
\label{tab:UpsampledBalAccLUAD}
\end{table*}

\begin{figure}[t!] 
    \centering
    \includegraphics[width=130pt]{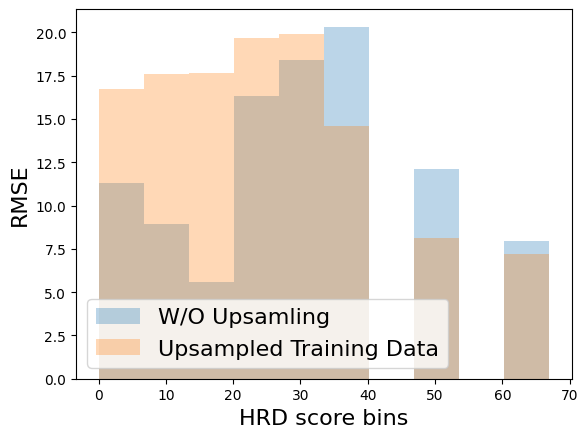}
    \caption{Binned RMSE of attMIL architecture trained with UNI feature extractor with (orange) and without (blue) upsampling the training data}
    \label{fig:binned_RMSE}
\end{figure}

Accurate prediction of underrepresented high-HRD patients is of clinical importance. When we compare the root mean squared error (RMSE) before and after upsampling, there's noticeable decrease in RMSE of underrepresented high-HRD patients and increase in RMSE of well-represented HRD values. This indicates that upsampling encourages the model to better capture rare HRD values while slightly compromising performance for common HRD levels. Fig.~\ref{fig:binned_RMSE} shows an representative example of such RMSE distribution shift using attMIL coupled UNI model and UCEC dataset.


In summary, upsampling improves balanced accuracy and recall for HRD+ patients, mitigating class imbalance. The decreased RMSE in high-HRD patients reflects an improved prediction of clinically relevant minority cases and represent a meaningful contribution to computational pathology.

\subsection{Ablation Experiments -- Impact of Bagsize and Sampling Strategies} 

\subsubsection{The Impact of Bagsize}
To assess the influence of bag size on MIL performance, an ablation study was conducted on the TCGA-BRCA cohort using five-fold cross-validation. All six feature extraction models were evaluated with the two MIL architectures, attMIL and SuRe Transformer, while varying the number of sampled image patches per slide. In addition to the bag sizes used in the main experiments, values ranging from 50 to 4000 patches were tested. Since we performed the k-Menas clustering with k=50, a bag size of 50 corresponds to selecting a single representative patch from each cluster when using the cluster size-weighted sampling strategy.

\begin{figure}[b!] 
    \centering
    \includegraphics[width=\linewidth]{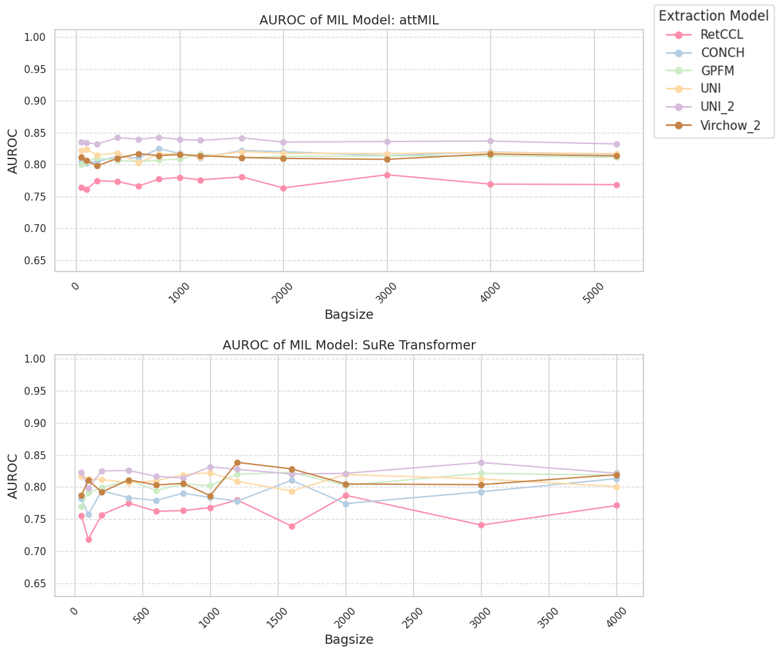}
    \caption{AUROC with attMIL or SuRe transformer architecture on TCGA-BRCA across different bagsizes}
    \label{fig:BagsizeKather}
\end{figure}

Across all configurations, no strong or consistent relationship between bag size and predictive performance was observed (Figure.~\ref{fig:BagsizeKather}). For the attMIL architecture, median AUROC (Figure.~\ref{fig:BagsizeKather}) and RMSE values (data not shown) remained largely stable as bag size increased, indicating that providing additional patches did not translate into systematic performance gains. This observation was confirmed quantitatively by weak correlations coefficients between bag size and performance metrics (0.2057 for AUROC and -0.1487 for RMSE), suggesting at most a marginal positive effect of larger bags. In contrast, the SuRe Transformer exhibited a slightly stronger dependency on bag size. Correlation coefficients of 0.3022 for AUROC and -0.3289 for RMSE indicate a weak but more noticeable improvement with increasing bag size. This trend is consistent with the design of the SuRe Transformer, whose radial decay self-attention mechanism emphasizes interactions between spatially proximal patches. Larger bags increase the probability that relevant local neighborhoods are adequately represented in the sampled set, allowing the attention mechanism to operate at finer spatial resolution \cite{SuReTransformer}


Overall, the ablation study suggests that bag size is not a dominant factor for HRD prediction performance in attention-based MIL models, while transformer-based aggregation may benefit modestly from increased context. 


\subsubsection{The Impact of Different Sampling Strategies}

To assess the impact of instance selection, we conducted an ablation study comparing different sampling strategies on the TCGA-BRCA cohort. Experiments were performed using five-fold cross-validation with the attMIL architecture across all six feature extractors and a range of bag sizes from 50 to 4000. To isolate the effect of sampling, performance was summarized as the mean of median AUROC values across six feature extractors.

\begin{figure}[] 
    \centering
    \includegraphics[width=180pt]{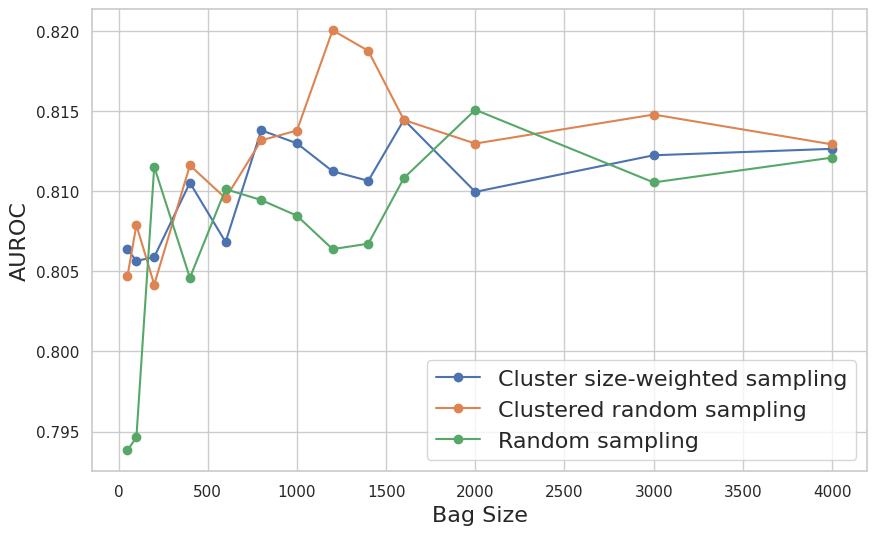}
    \caption{Mean AUROC for different sampling strategies}
    \label{fig:ablation_sampling}
\end{figure}

Overall, the performance differences between strategies were modest; however, their trajectories across bag sizes reveal clear trends (Fig.~\ref{fig:ablation_sampling}). Cluster-based sampling outperformed pure random sampling, with the advantage being most pronounced for small bag sizes (50 and 100), where random sampling exhibited substantially lower AUROC values. This indicates that enforcing cluster-level coverage leads to more informative bag representations.

Comparing the two cluster-based approaches, clustered random sampling achieved slightly higher average performance across all bagsizes than cluster-size weighted sampling. This suggests that sampling strictly proportional to cluster size is not optimal and may overemphasize large but less informative clusters. However the result distribution is very volatile, as observed with the unusually strong performance of random sampling at a bagsizes of 200 and 2000. To further validate the findings of this experiment additional investigations would be helpful.\\

In summary, the ablation experiments support the benefit of clustering-based instance selection for MIL-based HRD prediction, and effective patch selection and representation learning are more critical than the sheer number of sampled instances. Incorporating HRD-tailored clustering mechanisms that leverage domain knowledge to emphasize the most relevant image regions, represents a promising direction for future work.

\section{Conclusion}
Our results demonstrate the superior performance of large-scale foundation models for HRD prediction. Among the evaluated approaches, UNI-2 and Virchow-2, as the largest-scale models, consistently emerged as the strongest feature extractors across datasets and evaluation settings. In addition, addressing class imbalance through upsampling improves sensitivity to rare high-HRD cases, emphasizing the importance of optimization strategies that balance overall performance with clinically relevant detection. Finally, our findings highlight sophisticated instance selection strategies as a promising research direction, with clustering-based sampling yielding more robust performance than random selection, particularly under constrained bag sizes.

\bibliographystyle{named}
\bibliography{ijcai26}

\end{document}